\pdfoutput=1 
\documentclass[twoside,11pt]{article}
\usepackage{natbib}

\usepackage{multirow}

\usepackage{paper-style} 
\usepackage{subcaption,amsmath,booktabs,tabularx}
\usepackage{graphicx}
\usepackage{rotating}
\usepackage{pifont,mdframed}
\usepackage[]{algorithm2e}
\usepackage{xcolor}
\usepackage{hyperref}



\newcommand{\seVtwoRows}{507,805}

\newcommand{\seCoLoveGivenJoy}{6.1}
\newcommand{\seCoJoyGivenLove}{18.3}

\definecolor{goemotions}{HTML}{D62728}      
\definecolor{meld}{HTML}{2CA02C}            
\definecolor{crowdflower}{HTML}{9467BD}     
\definecolor{semeval}{HTML}{8C564B}         
\definecolor{emotion}{HTML}{E377C2}         

\usepackage{listings}
\title{The  SuperEmotion dataset}  
\author{Enric Junqu\'e de Fortuny\\
	Decision Sciences\\
	SKK GSB, Sungkyunkwan University, Seoul, South Korea\\
	enric@skku.edu
}
\nipsfinalcopy

\begin{document}
	\maketitle

\begin{abstract}
	Despite the wide-scale usage and development of emotion classification datasets in NLP, the field lacks a standardized, large-scale resource that follows a psychologically grounded taxonomy. Existing datasets either use inconsistent emotion categories, suffer from limited sample size, or focus on specific domains. The  SuperEmotion dataset addresses this gap by harmonizing diverse text sources into a unified framework based on Shaver's empirically validated emotion taxonomy, enabling more consistent cross-domain emotion recognition research. This report describes version 2 of the dataset. We also report baseline results for a fine-tuned ModernBERT encoder.
\end{abstract}
	
\section{Introduction}
\lettrine[lines=3]{T}{his} report describes the SuperEmotion dataset, the world's largest Shaver-compliant emotion dataset for natural language processing. We developed the  SuperEmotion dataset by aggregating multiple existing emotion datasets and remapping categories into Shaver's primary emotion classes. The dataset encompasses \seVtwoRows{} samples labeled across the primary emotions: \textit{joy, sadness, anger, fear, love}, and \textit{surprise}, as well as a \textit{neutral} category. Each of its sources uses its own label set based on text drawn from Reddit comments, tweets and television dialogue, yet all share the same seven categories.
\vspace{0pt plus 14pt} 

By consolidating these resources, the  SuperEmotion dataset addresses class imbalances and provides a more diverse and extensive foundation for training robust emotion classification models following a well-understood taxonomy. Section~\ref{sec:construction} describes the sources and how we built the dataset, Section~\ref{sec:baselines} gives baseline results, and Section~\ref{sec:access} explains how to access the dataset and its limitations.

\section{Dataset Construction}
\label{sec:construction}

\subsection{Dataset Composition}

The  SuperEmotion dataset is constructed by aggregating five publicly available emotion classification datasets:

\begin{itemize}
	\item \textbf{MELD} \citep{poria_meld_2019}: A multimodal dataset for emotion recognition in conversations. The text comes from scripts and transcribed dialogues from the TV show \textit{Friends}, capturing multi-party spoken interactions with speaker context.
	
	\item \textbf{GoEmotions} \citep{mohammad2010emotions}: A large-scale dataset of carefully filtered English Reddit comments annotated with 27 fine-grained emotions plus neutrality. It is well-suited for studying subtle affect in user-generated social media content.
	
	\item \textbf{CARER} \citep{saravia_carer_2018}: A Twitter-based dataset developed for context-aware emotion recognition, containing tweets labeled across six emotion categories. The short, informal text reflects real-world online communication.
	
	\item \textbf{SemEval} \citep{mohammad_semeval-2018_2018}: A dataset from the Semantic Evaluation series, particularly Task 1 on Affect in Tweets. It contains tweets annotated for emotional intensity and valence, providing rich insight into affective language in short-form, real-time social media.
	
	\item \textbf{CrowdFlower} \citep{van_pelt_designing_2012}: A crowdsourced dataset of tweets labeled with emotion categories such as sadness, joy, and anger. The data was used to evaluate how well laypeople agree on emotion labels in noisy, informal text.
\end{itemize}

\noindent Table~\ref{tab:dataset_sources} summarizes their contribution to each split.

\begin{table*}[t]
	\centering
	\begin{tabular*}{\textwidth}{@{}l@{\extracolsep{\fill}}rrrrr@{}}
		\toprule
		\textbf{Dataset} & \textbf{Classes} & \textbf{Original release} & \textbf{Train} & \textbf{Val.} & \textbf{Test} \\
		\midrule
		MELD & 7 & 13,708 & 8,901 & 966 & 2,256 \\
CARER & 6 & 416,809 & 318,668 & 37,723 & 37,428 \\
GoEmotions & 28 & 54,263 & 41,840 & 5,218 & 5,208 \\
CrowdFlower & 13 & 39,998 & 31,210 & 3,896 & 3,893 \\
SemEval & 11 & 10,690 & 6,567 & 866 & 3,165 \\
\midrule
\textbf{SuperEmotion} & 7 & 535,468 & 407,186 & 48,669 & 51,950 \\
 
		\bottomrule
	\end{tabular*}
	\caption{The source datasets: emotion classes, size of the original release, and texts in each SuperEmotion split (version 2, after label mapping and deduplication). We created the splits of CARER and CrowdFlower.}
	\label{tab:dataset_sources}
\end{table*}

\subsection{Preprocessing and Quality Control}
To ensure consistency across diverse source formats, we applied several preprocessing steps:

\begin{enumerate}
	\item \textbf{Text Normalization}: We standardized text formatting by removing excessive whitespace, normalizing Unicode characters, and ensuring consistent punctuation placement.
	\item \textbf{Deduplication}: Every text occurs once, in one split, to prevent test set leakage and avoid biasing the dataset toward redundant patterns. A text that occurs in more than one split is kept in the first of train, validation and test. Copies within a split are merged, keeping each label that at least half of the copies carry.
	\item \textbf{Data Splits}: For datasets without predefined splits, we created stratified train/validation/test partitions (80\%/10\%/10\%) to preserve label distribution across splits.
	\item \textbf{Metadata Preservation}: While harmonizing the emotion taxonomy, we retained source information for each example, enabling analysis of domain-specific patterns and potential biases.		
\end{enumerate}

\subsection{Shaver's Taxonomy}

Emotion classification has been approached in a variety of ways in psychology, linguistics, and affective computing. Some models, such as Ekman's framework \citep{eckman_universal_1972}, define a small set of discrete universal emotions (e.g., anger, fear, joy, sadness, surprise, disgust), primarily grounded in facial expressions. Others, like Russell's circumplex model \citep{russell_core_1999}, represent emotions in a continuous space defined by valence and arousal dimensions. Plutchik's wheel of emotions \citep{plutchik_general_1980,plutchik_nature_2001} combines discrete emotions with intensity scaling and mixing of different primary emotions.

We adopt the taxonomy proposed by \citet{shaver_emotion_1987}, which provides a psychologically grounded, hierarchical classification of emotions based on empirical clustering of 135 emotion terms. Through free-listing, sorting, and similarity judgments, Shaver et al. identified six basic-level emotions (\textit{love, joy, anger, sadness, fear,} and \textit{surprise}\footnote{Note: We include \textit{surprise} even though Shaver gave it less importance, as it appears in all datasets and is clearly of interest to the NLP community. 
}), which serve as cognitively salient and linguistically frequent prototypes in English.

We selected Shaver's taxonomy for two reasons. First, it aligns well with the input modality of most NLP systems, as it is lexically grounded, built from natural language emotion terms. Second, it balances granularity and coverage, capturing sufficient emotional nuance for robust modeling while remaining tractable for supervised learning and being robust to potential expansions of the dataset.

\subsection{Label Harmonization}
To align the heterogeneous label sets of the source datasets, we mapped each source label to one of the six core categories, using Shaver's original definitions and keywords \citep{shaver_emotion_1987}. We also introduce a \textit{neutral} category for texts with low or absent emotional valence. Table~\ref{tab:shaver-report} gives the complete mapping.

For ambiguous labels we prioritized semantic similarity to Shaver's prototypical emotion terms. For instance, \textit{optimism} could reasonably map to \textit{joy} or be seen as a distinct anticipatory state, but we assigned it to \textit{joy} based on its positive valence and association with pleasant feelings. Similarly, \textit{confusion} was mapped to \textit{surprise} rather than \textit{fear} based on its cognitive rather than threat-oriented nature.

Three labels lacked a clear conceptual correspondence with Shaver's taxonomy: \textit{anticipation}, \textit{curiosity} and \textit{empty}. Rather than force them into potentially inappropriate categories, we dropped them. A text keeps its other labels, but a text left without any label is removed.

\begin{table*}[t]
	\centering
	\footnotesize
	\begin{tabularx}{\textwidth}{@{}l>{\raggedright\arraybackslash\hsize=0.7\hsize}X>{\raggedright\arraybackslash\hsize=0.55\hsize}X>{\raggedright\arraybackslash\hsize=2.05\hsize}X>{\raggedright\arraybackslash\hsize=0.95\hsize}X>{\raggedright\arraybackslash\hsize=0.75\hsize}X@{}}
		\toprule
		\textbf{Emotion} & \textbf{MELD} & \textbf{CARER} & \textbf{GoEmotions} & \textbf{CrowdFlower} & \textbf{SemEval} \\
		\midrule
		\textit{anger} & anger, disgust & anger & anger, annoyance, disapproval, disgust & anger, hate & anger, disgust \\
		\textit{fear} & fear & fear & fear, nervousness & worry & fear \\
		\textit{joy} & joy & joy & amusement, excitement, joy, optimism, pride, relief & enthusiasm, fun, happiness, relief & joy, optimism \\
		\textit{love} & & love & admiration, approval, caring, desire, gratitude, love & love & love, trust \\
		\textit{sadness} & sadness & sadness & disappointment, embarrassment, grief, remorse, sadness & sadness & pessimism, sadness \\
		\textit{surprise} & surprise & surprise & confusion, realization, surprise & surprise & surprise \\
		\midrule
		\textit{neutral} & neutral & & neutral & boredom, neutral & \\
		Dropped & & & curiosity & empty & anticipation \\
		\bottomrule
	\end{tabularx}
	\caption{Mapping of the source labels to Shaver's categories, per source dataset.}
	\label{tab:shaver-report}
\end{table*}

\begin{table*}[t]
	\centering
	\small
	\begin{tabular*}{\textwidth}{@{}l@{\extracolsep{\fill}}rrrrrrrrr@{}}
		\toprule
		\textbf{Source} & \textit{neutral} & \textit{surprise} & \textit{fear} & \textit{sadness} & \textit{joy} & \textit{anger} & \textit{love} & \textbf{Dropped} & \textbf{Total} \\
		\midrule
		MELD & 5,662 & 1,311 & 344 & 951 & 2,042 & 1,854 & 0 & 0 & 12,164 \\
CARER & 2 & 14,076 & 46,284 & 120,139 & 138,846 & 56,449 & 32,809 & 0 & 408,605 \\
GoEmotions & 17,723 & 4,287 & 926 & 4,021 & 7,615 & 7,818 & 15,423 & 2,723 & 60,536 \\
Crowdflower & 8,782 & 2,182 & 8,442 & 5,155 & 9,237 & 1,433 & 3,792 & 827 & 39,850 \\
SemEval & 0 & 566 & 1,848 & 3,607 & 5,065 & 4,779 & 1,757 & 1,527 & 19,149 \\
\midrule
\textbf{Total} & 32,169 & 22,422 & 57,844 & 133,873 & 162,805 & 72,333 & 53,781 & 5,077 & 540,304 \\
 
		\bottomrule
	\end{tabular*}
	\caption{Labels per emotion category and source in SuperEmotion version 2, all splits together. The 41 texts found in two sources are counted with the first.}
	\label{tab:emotion_distribution}
\end{table*}

Table~\ref{tab:emotion_distribution} summarizes the distribution of emotion labels across all source datasets after mapping to the unified taxonomy. Note that counts may exceed those in Table~\ref{tab:dataset_sources} due to the multi-label nature of the task, where a single text sample can be annotated with multiple emotions, and column ``Dropped'' counts the labels removed in the mapping.

\begin{figure}[t]
	\begin{minipage}[c]{0.58\textwidth}
		\centering
		\includegraphics[width=\linewidth]{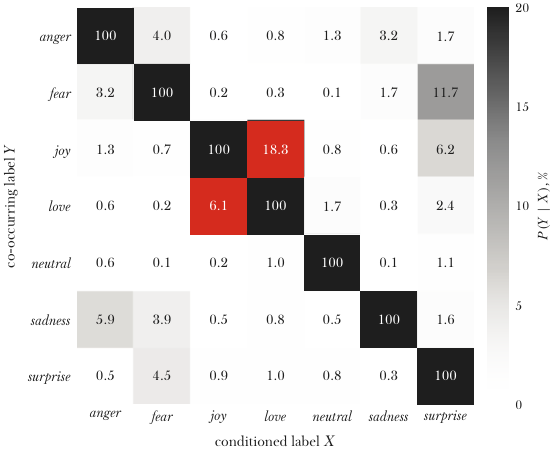}
		\caption{Label co-occurrence heatmap showing the percentage of samples annotated with emotion  $X$ (X-axis) that are also annotated with emotion $Y$ (Y-axis), denoted as $P(Y\,|\,X)=\frac{\#(X \cap Y)}{\#(X)}$. Diagonal values are always 100\%, as each annotation trivially co-occurs with itself.}
		\label{fig:label_cooccurrence}
	\end{minipage}\hfill
	\begin{minipage}[c]{0.38\textwidth}
		\centering
		\begin{tabular}{@{}lrccc@{}}
			\toprule
			& \textbf{n} & \textbf{Prec.} & \textbf{Rec.} & \textbf{F1} \\
			\midrule
			\input{tables/se_results.tex} 
			\bottomrule
		\end{tabular}
		\captionof{table}{ModernBERT-base on SuperEmotion (version 2), threshold 0.5 for all labels: macro averages per split and scores per emotion on the test split. The model is trained once on all seven labels.}
		\label{tab:baselines}
	\end{minipage}
\end{figure}

Figure~\ref{fig:label_cooccurrence} visualizes the overlap between emotion categories, showing how frequently different emotions co-occur within the same instance. More precisely, it visualizes the conditional probability of observing emotion $Y$ given that emotion $X$ is present, expressed as a percentage. The asymmetric nature of the matrix reflects that $P(Y|X) \neq P(X|Y)$ for most emotion pairs. For example, \seCoLoveGivenJoy\% of texts annotated as \textit{joy} also contain \textit{love}, while \seCoJoyGivenLove\% of texts labeled as \textit{love} also contain \textit{joy}.

\section{Baseline}
\label{sec:baselines}
To give users a reference point, we fine-tune ModernBERT-base \citep{warner_smarter_2024} on the training split of version 2. The recipe is a linear head with a sigmoid per label and binary cross-entropy loss, AdamW \citep{loshchilov_decoupled_2017} with a learning rate of 2e-5, a batch size of 16, a weight decay of 0.01, a cosine schedule, inputs truncated to 128 tokens, and up to five epochs with evaluation after each epoch and early stopping after three epochs without improvement. The checkpoint with the best validation macro-F1 is kept.

Table~\ref{tab:baselines} gives precision, recall and F1 at a threshold of 0.5, per split and per emotion. Macro-F1 is the mean of the per-label F1 scores. Thresholds tuned per label on validation do not help (test macro-F1 0.836). The model is available on Hugging Face at \url{https://huggingface.co/cirimus/modernbert-base-emotions} and can be tried in the browser at \url{https://huggingface.co/spaces/cirimus/emotion-detection}.

\section{Final Considerations}
\subsection{Data Accessibility}
\label{sec:access}

The dataset is publicly available on Hugging Face at the following URL: 	\url{https://huggingface.co/datasets/cirimus/super-emotion}. Alternatively, users can download the dataset using the Hugging Face \texttt{datasets} library in Python:

\begin{lstlisting}[language=Python]
from datasets import load_dataset
dataset = load_dataset("cirimus/super-emotion")
\end{lstlisting}

\noindent When the dataset is updated, we will update this versioned repository while keeping the old data available as a tagged version. The dataset described in this document is version 2. Version 1 remains available in the same repository as the tag \texttt{v1} (\texttt{revision="v1"} in \texttt{load\_dataset}).

\subsection{Limitations and Ethical Considerations}
While the  SuperEmotion dataset offers advantages in scale and emotional coverage, several limitations should be acknowledged:

\begin{enumerate}
	\item \textbf{Dataset Biases}: The aggregation inherits biases from source datasets, including potential cultural and demographic skews in emotion expression and annotation. Moreover, the dataset introduces a new sampling bias due to CARER’s dominance in size. Researchers may therefore want to stratify across datasets to mitigate this problem.
	\item \textbf{Contextual Limitations}: Many samples lack conversational context that might influence emotion interpretation.
	\item \textbf{Annotation Quality}: Source datasets employed different annotation methodologies and annotator populations, potentially introducing inconsistencies in label quality.
	\item \textbf{Privacy Considerations}: Though all datasets excluded personally identifiable information, users should remain cautious when deploying models trained on this data in applications involving sensitive contexts.
\end{enumerate}

\noindent We encourage researchers to consider these limitations when developing emotion recognition systems and to supplement with domain-specific data when appropriate.

\section{Conclusion}
By creating the  SuperEmotion dataset, we contribute to emotion recognition research by harmonizing multiple existing datasets into a consistent taxonomy based on Shaver's psychological framework. By addressing class imbalances, providing a diverse text collection and reducing taxonomical inconsistencies, this resource enables more robust emotion classification models.

The harmonized labels, clear documentation, and easy accessibility through Hugging Face facilitate immediate application in natural language processing tasks. We anticipate this dataset will support advances in affective computing, human-computer interaction, and sentiment analysis. Future work may expand this collection with additional data, especially for the secondary emotions in Shaver's hierarchy.

\bibliography{references}
	
\end{document}